\documentclass[conference,a4paper]{ieeetran}
\usepackage[latin1]{inputenc}
\usepackage{subeqnarray}
\usepackage{amsfonts}
\usepackage{amsmath}
\usepackage{amssymb}
\usepackage{graphicx}
\usepackage{color,flushend}
\usepackage[bookmarks=false]{hyperref}

\usepackage{fancyheadings}
\title{Skin Lesion Segmentation: U-Nets versus Clustering
}

\author{\authorblockN{Bill S. Lin\authorrefmark{1}, Kevin Michael\authorrefmark{2}, Shivam Kalra\authorrefmark{3}, H.R. Tizhoosh\authorrefmark{3}
}
\authorblockA{\authorrefmark{1} Department of Mechanical and Mechatronics, University of Waterloo, Waterloo, Canada}
\authorblockA{\authorrefmark{2} Department of Systems Design Engineering, University of Waterloo, Waterloo, Canada}
\authorblockA{\authorrefmark{3}  KIMIA Lab, University of Waterloo, Waterloo, Canada}
}

\begin{document}
\maketitle
\begin{abstract}
Many automatic skin lesion diagnosis systems use segmentation as a preprocessing
step to diagnose skin conditions because skin lesion shape, border irregularity,
and size can influence the likelihood of malignancy. This paper presents,
examines and compares two different approaches to skin lesion segmentation. The
first approach uses U-Nets and introduces a histogram equalization based
preprocessing step. The second approach is a C-Means clustering based approach that is much simpler to implement and faster to execute. The Jaccard Index between the algorithm output and hand segmented images by dermatologists is used to evaluate the proposed algorithms. While many recently proposed deep neural networks to segment skin lesions require a significant amount of computational power for training (i.e., computer with GPUs), the main objective of this paper is to
present methods that can be used with only a CPU. This severely limits, for
example, the number of training instances that can be presented to the U-Net. Comparing the two proposed algorithms, U-Nets achieved a
significantly higher Jaccard Index compared to the clustering approach.
Moreover, using the histogram equalization for preprocessing step significantly
improved the U-Net segmentation results.
\end{abstract}


\section{Introduction}
With between 2 and 3 million cases occurring globally each year, skin cancer is
the most common cancer worldwide \cite{1,2}. Since skin cancer occurs at the
surface of the skin, melanomas, one of the world's most deadly cancer, can be
diagnosed with visual inspection by a dermatologist \cite{1,3}. To improve the
accuracy, dermatologists use a dermatoscope, which eliminates some of the
surface reflection and enhances the deeper layers of skin \cite{4}.
Dermatologists often look for five specific signs when classifying melanomas.
These include skin lesion asymmetry, irregular borders, uneven distribution of
color, diameter, and evolution of moles over time \cite{5}. Standard skin cancer
computer-aided diagnosis systems include five important steps: acquisition,
preprocessing, segmentation, feature extraction, and finally, classification
\cite{6}. Since the shape and border of the skin lesion is extremely important
in the diagnosis, automatic diagnosis systems must be able to accurately
identify and segment the skin lesion in the image.

Automatic skin lesion segmentation contains many different challenges. The
lesion color, texture, and size can all vary considerably for different
patients. On top of this, medical gauzes, hair, veins, and light reflections in
the image make it even more difficult to identify the skin lesion. Finally, in
the areas surrounding a skin lesion, there are often sub-regions that are darker
than others. This makes it difficult to consistently identify which sub-regions
are part of the lesion and which sub-regions are just part of the surrounding
skin. Finally, lack of large labeled datasets (segmented by dermatologists)
makes training sophisticated networks rather difficult.

This paper presents and compares multiple approaches to automatic segmentation
of dermoscopic skin lesion images. Firstly, we present some different
segmentation approaches using U-nets including one with a novel histogram-based
preprocessing technique. After this, the U-net approach is compared with a much
simpler clustering algorithm. While U-nets and other deep networks often require
a lot of training and computational power, the approaches presented in this
paper are designed such that they do not require GPU power. All training and
testing in this paper have been done with a single CPU.

The paper is organized as follows: Section~\ref{sec:background} briefly reviews the literature of existing techniques for skin lesion segmentation.
Section~\ref{sec:image_data} describes the proposed dataset used to benchmark proposed methods. Section~\ref{sec:methodology} explains the proposed methods; results and conclusion are discussed in Section~\ref{sec:results_dicsussion} and
Section~\ref{sec:conclusion}, respectively.

\section{Background Review} \label{sec:background}
Machine learning schemes like reinforcement learning and neural networks have
been used for segmentation of medical images \cite{19,20,21,22,23}.
Since skin lesion segmentation is frequently used as an important first step in
many automated skin lesion recognition systems, there have been many different
approaches. A summary of several skin lesion segmentation methods by Celebi et
al. \cite{7} reviews many recent techniques proposed in literature. Frequently,
proposed lesion segmentation techniques include a preprocessing step which
removes unwanted artifacts such as hair or light reflections. Another very
common preprocessing step involves performing color space transformation in
order to make segmentation easier. Common techniques that have recently been
proposed in the literature include, among others, histogram thresholding,
clustering and active contours. Most recently, deep neural networks have led to
many breakthroughs, producing excellent results in image segmentation \cite{8}.

Color space transformations can help facilitate skin lesion detection. For
instance, Garnavi et al. \cite{9}, take advantage of RGB, HSV, HSI, CIE-XYZ,
CIE-LAB, and YCbCr color spaces and use a thresholding approach to separate the
lesion from the background skin. Similarly, Yuan et al. \cite{10} use RGB, HSV,
and CIE-LAB color spaces as input channels to a fully
convolutional-deconvolutional neural network in order to segment the images.

Hair removal is a very common preprocessing step used to facilitate skin lesion
segmentation. The presence of dark hairs on the skin often significantly
degrades the performance of most automatic segmentation algorithms. For example,
one very popular software, Dullrazor \cite{11}, identifies the dark hair
locations using morphological operations. Afterwards, the hair pixels are
replaced with nearby skin pixels and the resulting image is smoothed in order to
remove the hair artifacts. Many recent techniques uses similar morphology or
diffusion methods to remove hair \cite{6, 9, 12}.

Once the image has been preprocessed, many different methods have been proposed
for segmentation. Since most skin lesions are darker than the surrounding skin,
thresholding is a very commonly used technique. However, further image
enhancement algorithms need to be performed in order to make thresholding
effective. For example, Fan et al. \cite{6} propose first enhancing the skin
lesion image using color and brightness saliency maps before applying a modified
Otsu threshold method. Similarly, Garnavi et al. \cite{9} use a thresholding
based approach on 25 color channels from a variety of different color spaces in
order to segment the images. Often with histogram thresholding, a connected
components analysis needs to be performed after thresholding in order to remove
medical gauzes, pen marks, or other artifacts that have been wrongly classified
by thresholding.

Clustering-based methods are commonly used as well. For instance, Schmid-Saugeon
et al. \cite{12} propose using a modified Fuzzy C-Means clustering algorithm on
the two largest principle components of the image in the LUV color space. The
number of clusters for the Fuzzy C-Mean algorithm is determined by examining the
histogram computed using the two principal components of the image. Melli et al.
\cite{13} propose using mean shift clustering to get a set of redundant classes
which are then merged into skin and lesion regions.

More recently, convolutional neural networks and other deep artificial neural
networks are proposed. Commonly, the image is first preprocessed to remove
artifacts. Afterwards, a neural network is used to identify the lesion. For
instance, in a work proposed by Jafari et al. \cite{14}, the image is first
preprocessed using a guided filter before a CNN uses both local and global
information to output a label for each pixel.

\section{Image Data} \label{sec:image_data}
The 2017 IEEE International Symposium on Biomedical Imaging (ISBI) organized a
skin lesion analysis challenge for melanoma detection, ISIC 2017 (International
Skin Imaging Collaboration). The challenge released a public set of 2000 skin
lesion images that have been segmented and classified by dermatologists. There
is also a validation and a test set that is not publicly available for the
competition. Lesion segmentation is one of three parts of the competition. For
evaluation, the Jaccard Index is used to rank the algorithms. This year, the
vast majority of the participants, including the top submissions, used a deep
neural networks to segment the images. For example, Yuan et al. [10] augmented
the 2000 images using rotation, flipping, rotating, and scaling and trained a
convolutional-deconvolutional neural network with 500 epochs. Clustering was
also used in some submissions, although the performance of clustering algorithms
were generally worse compared with that of the deep artificial networks. A
variety of different evaluation metrics have been used in the literature. This
includes the XOR measure, and Specificity, Sensitivity, Precision, Recall, and
error probability \cite{7}. The ISBI 2017 challenge used the Jaccard Index in
order to evaluate methods. In order to compare the algorithms and methods
proposed in this paper, the average Jaccard Index is used to evaluate the
proposed method's performance.

The ISIC 2017 training set was used to both train and validate our approaches.
This publicly available dataset contains a total of 2000 images including 374
melanoma, 254 seborrheic keratosis and 1372 benign nevi skin lesion images in
JPG format. The images are in many different dimensions and resolutions. Each
skin lesion has also been segmented by a dermatologist; the 2000 binary masks
corresponding to each skin lesion image is provided in PNG format.

\begin{figure}[t!]
\begin{center}
\includegraphics[width=\columnwidth]{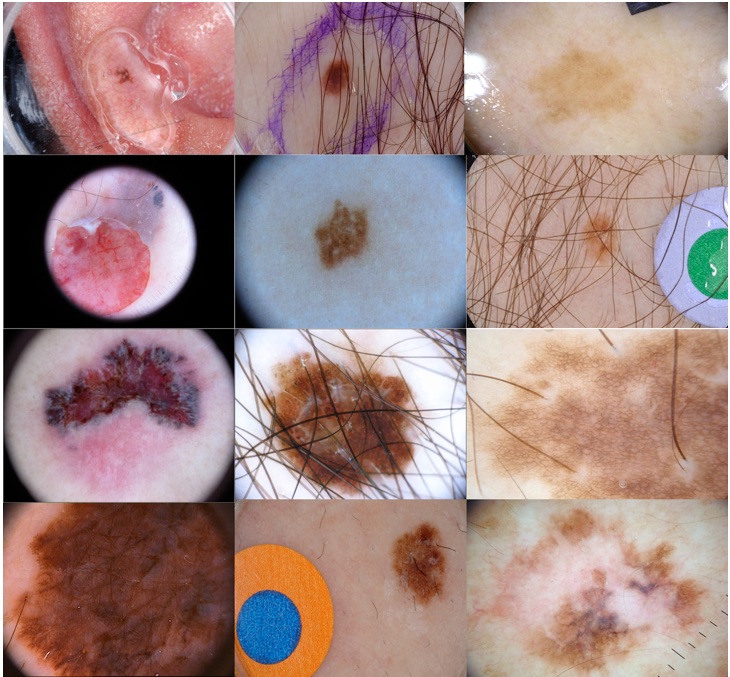}
\caption{Sample skin lesion images from public data set used for ISBI 2017 challenge.}
\label{fig:sample_images}
\end{center}
\end{figure}

Figure~\ref{fig:sample_images} shows a collection of sample images from the data
set. As can be seen from the images, there are many unwanted artifacts including
reflections, hair, black borders, pen marks, dermatoscope gel bubbles, medical
gauzes, and other instrument markings that make segmentation difficult.
Furthermore, there is a high variability in skin lesion shape, size, color and
location.
 
\section{Methodology} \label{sec:methodology}
The main objective of this paper is to develop different approaches that can
process skin lesion images, and produce a binary mask with 1 corresponding to
pixels that are part of the skin lesion, and 0 corresponding to pixels that are
not part of the skin lesion. The output format should be identical to the binary
masks segmented by the dermatologists in order to evaluate the performance of
the segmentation algorithms.

In order to compare the performance of algorithms with that of other submissions
in the ISBI challenge, the Jaccard Index, which was the challenge evaluation
metric, is used. Since all of the approaches in this paper are performed without
a GPU, some approaches, in particular, the ones involving a U-Net, take around
several hours (in our case 8 hours) to train for each validation fold. In order
for the validation to complete within a reasonable amount of time, 5 random
cross validation folds were performed with a 90\% training (1800 images) and
10\% validation (200 images) split. In this methodology section, we first
discuss the performance of U-Nets followed by clustering.

\subsection{U-Net Approach}
\begin{figure*}[h]
\begin{center}
\includegraphics[width=0.85\textwidth]{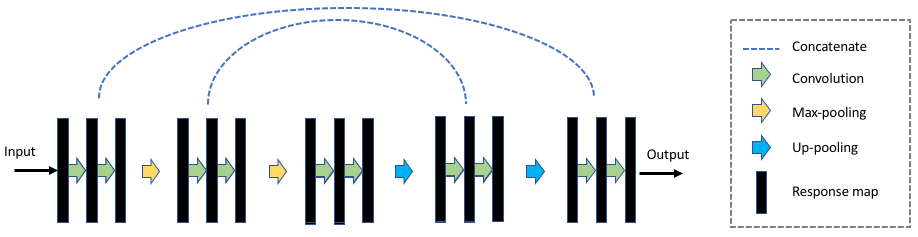}
\caption{Schematic illustration of U-Net according to \cite{16}.}
\label{fig:unet}
\end{center}
\end{figure*}

The U-Net architecture used in image segmentation uses a Python library for
U-Nets by Akeret et al. \cite{15} with the open source library TensorFlow. The
architecture is based on the network proposed by Ronneberger et al. \cite{16}
for biomedical image segmentation (Figure \ref{fig:unet}). As discussed in the
paper by Ronneberger et al., this network architecture can be trained with very
few images and is relatively fast to train and test.

Because the goal of this paper is to develop an algorithm that works without the
need of a GPU, the number of training iterations that can be done on the U-Net
is severely limited. Unlike the top ISBI submissions that used Neural Networks,
which augment the data to tens of thousands of images and set the number of
epochs to a few hundred \cite{10}, the proposed network was trained with a lot
less training iterations.

The 1800 images are first augmented to 7200 images by a combination of flips and
rotations. Afterwards, just over one epoch (10,000 training iterations) is
applied to train the network. This is roughly 1\% of the number of training
iterations used by the top submissions in the ISBI challenge.

\subsubsection{Architecture and Training}
Similar to convolutional neural networks, which repeatedly apply convolutions,
activation functions and downsampling, the U-Net architecture extends this by
including symmetric expansive layers which contain upsampling operators. The
proposed architecture contains 4 contracting steps and 4 expansion steps. Each
contraction step contains two unpadded $3\times 3$ convolutions each followed by
a ReLU activation function. Finally, a $2\times 2$ max pooling operation is used
for down sampling. In each down sampling step, the number of feature channels
doubles.

In each expansion step, two unpadded $3\times 3$ convolutions each followed by a
ReLU activation function is used. However, instead of the $2\times 2$ max
pooling, an up-convolution layer is added which halves the number of feature
channels. The features from the corresponding contraction layer are also
concatenated in the expansion layer.

At the end, a final layer of a $1\times 1$ convolution is used to map the
feature vector to a desired binary decision. A pixel-wise soft-max is applied in
order to determine the degree of which a certain pixel belongs to the skin
lesion.

The loss function used is the cross entropy between the prediction and the
ground truth. The Adam optimizer with a constant learning rate of $0.0002$ is
used to train the network. A dropout rate of $0.5$ is used.

\subsubsection{Pre- and Post-Processing}
Before the data is sent to the U-Net, a few pre-processing steps are performed.
Firstly, in order to increase the amount of contrast in the image, the image is
converted to HSI color space, and histogram equalization is applied to the
intensity (I) channel. After this, the image is converted back to RGB format.
The reason for this equalization is that the contrast between the skin lesions
(usually darker) is more noticeable compared to the surrounding skin (usually
brighter). On top of this, as a 4th channel, the original intensity (I) channel
with values normalized to between 0 and 1 is added. Finally, as a 5th channel, a
2-dimensional Gaussian centered at the middle of the image is generated. The
full-width half maximum of the 2D Guassian was set to be 125 pixels, which was
roughly where most of the skin lesions lie within the center from a visual
inspection of the training data.

To the authors' best knowledge, this is the first time that histogram
equalization and an extra Gaussian channel is used as an input to a U-Net. The
image needs to be rescaled and zero-padded so that the U-Net can perform
segmentation. Each image is rescaled such that the largest dimension is 250
pixels. The smaller dimension is filled with white padding so that the image
becomes $250\times 250$. Finally, in order to account for cropping due to the
unpadded convolutions, the image is white padded with a border of 46 pixels so
that the final size is $342\times 342$ pixels.

After training, some post-processing steps are applied to the testing set in
order to improve the result. Firstly, a morphology operation (using the
``scipy'' \emph{binary\_fill\_holes} function) is applied to the image in order
to fill the holes in the segmented image. After this, in order to find the best
threshold to set for the U-Net output, a gradient descent is applied to the
threshold using in order to determine the best threshold.

\subsubsection{U-Net Algorithm} 
In order to test the effect of the pre-processing and post-processing, two
different approaches are compared. One with the pre-/post-processing and one
without pre-/post-processing. The exact same U-Net architecture, which is
described in an earlier section, is used for both cases.

\begin{figure}[h]
\begin{center}
\includegraphics[width=0.85\columnwidth]{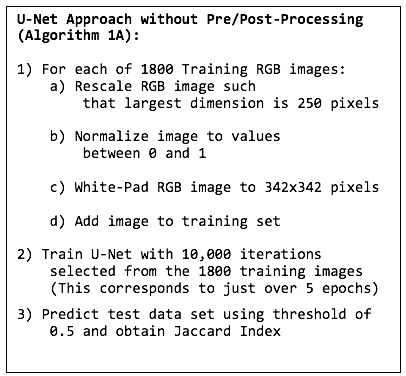}
\caption{U-Net Algorithm without Pre-/Post-Processing (Algorithm 1A)}
\label{fig:Alg1a}
\end{center}
\end{figure}

\begin{figure}[hb!]
\begin{center}
\includegraphics[width=0.85\columnwidth]{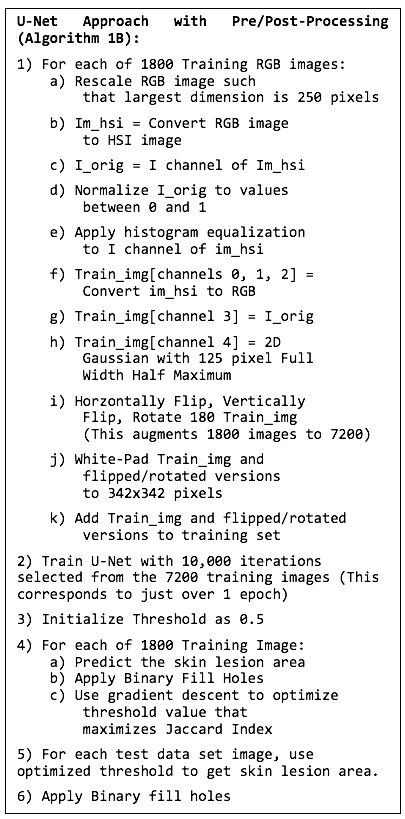}
\caption{U-Net Approach with Pre/Post-Processing (Algorithm 1B)}
\label{fig:Alg1b}
\end{center}
\end{figure}

The pseudo-code in Figure~\ref{fig:Alg1a} shows the algorithm without
pre-/post-processing. The pseudo-code in Figure \ref{fig:Alg1b} shows the
algorithm with pre/post-processing. The results obtained from the two approaches
are discussed in a later section.

\subsection{Clustering Algorithm}
This approach leverages an unsupervised clustering technique to analyze the
image and segment the lesion of interest. It uses Fuzzy C-means clustering in
order to split the image up into a number of distinct regions of interest. It
then leverages k-means clustering to group the clusters based on color features.
Ultimately it selects the group of clusters with the darkest color features and
identifies it to be the mole.

\subsubsection{Pre-Processing}
\begin{figure}[h]
\begin{center}
\includegraphics[width=0.85\columnwidth]{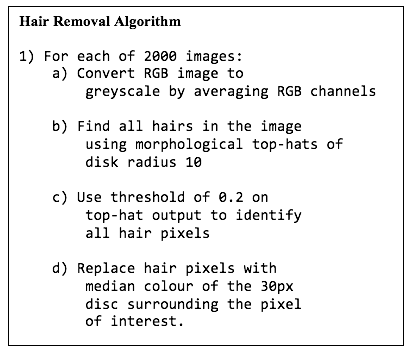}
\caption{Hair Removal Algorithm}
\label{fig:HairAlg}
\end{center}
\end{figure}

When conducting clustering to the images, there are two major sources of
disturbance within the image. Firstly, the presence of hair on-top of the skin
causes discrepancies in the algorithms ability to cluster, therefore strands of
hair were removed in a pre-processing step which utilizes a morphological
top-hat and thresholding technique. This technique is outlined in pseudo code in
Figure~\ref{fig:HairAlg}.

\begin{figure}[h]
\begin{center}
\includegraphics[width=0.85\columnwidth]{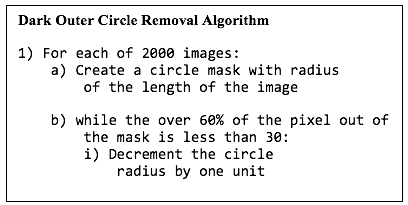}
\caption{Dark outer Circle Removal Algorithms}
\label{fig:outercircle}
\end{center}
\end{figure}

Secondly, a subset of the images were captured using imaging devices with a
circular field of view. Resultant images captured using this type of device were
circular in nature with black pixels filling the outer boundaries of the circle.
These pixels were removed from the area of analysis and the algorithm for doing
so can be seen in Figure~\ref{fig:outercircle}.

\subsubsection{Clustering Algorithm Summary}
\begin{figure}[h]
\begin{center}
\includegraphics[width=0.85\columnwidth]{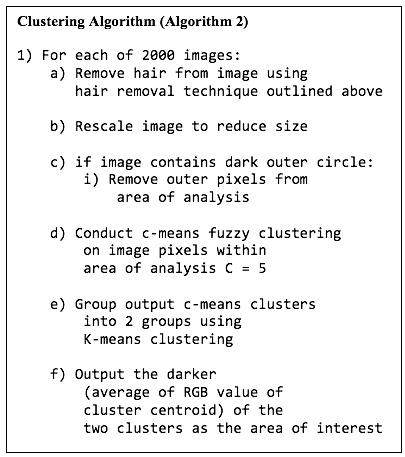}
\caption{Clustering Algorithm}
\label{fig:clustering}
\end{center}
\end{figure}

The algorithm outlined in the above sections can be found summarized in the
pseudo-code in Figure~\ref{fig:clustering} and includes the pre-processing steps
of hair removal and outer-circle removal. This algorithm will be referenced as
Algorithm 2 for future sections. Experimentally, when $C = 5$, at least one of
the clusters is the mole itself. For this reason, $C = 5$ was used for the
algorithm.

\section{Results and Discussions} \label{sec:results_dicsussion}
As discussed in the Methodology section, the segmented images are assessed using
the Jaccard Index through 5 random folds validation with 90\%-10\% train-test
split. For the clustering method, since it is completely unsupervised, the
average Jaccard Index for the entire data set is also calculated.

\begin{figure*}[h]
\begin{center}
\includegraphics[width=0.7\textwidth]{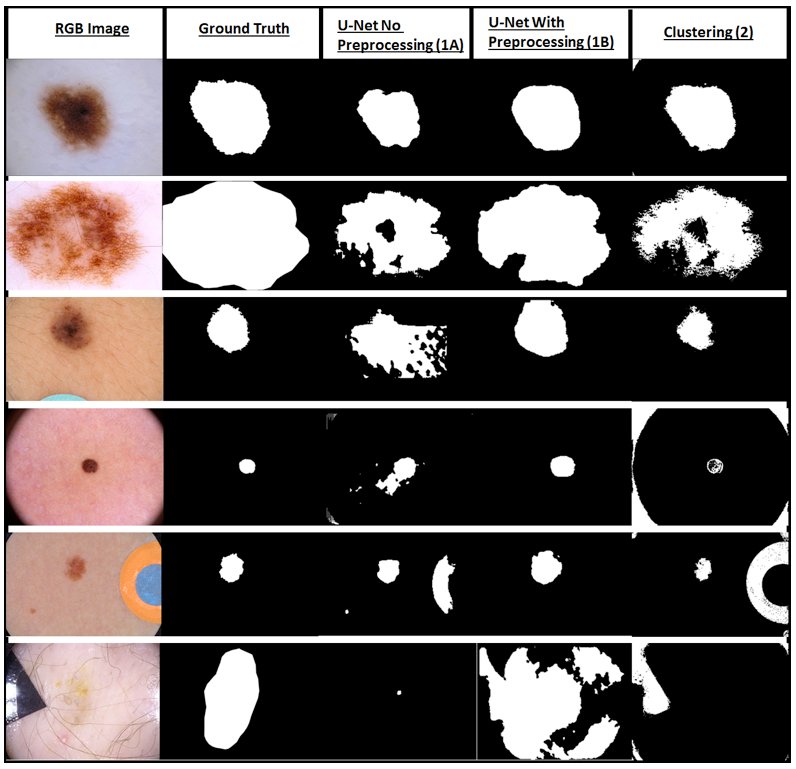}
\caption{Sample skin lesion segmentations. Top three rows: Typical examples of most skin lesion images; 3rd and 4rd row: Color variation in skin lesions; 5th row: Medical gauze example.}
\label{fig:results}
\end{center}
\end{figure*}

The Jaccard index and dice values for 5 different folds are presented in Table
\ref{tab:table}. Figure \ref{fig:results} shows sample segmentations.

\begin{table*}[h]
\centering
\caption{Jaccard index (and Dice coefficient in parentheses) for 5 different folds ($m$ is the average and $\sigma$ the standard deviation). }
\label{tab:table}
\begin{tabular}{l | lll}
Fold             & Algorithm 1A   & Algorithm 1B  & Algorithm 2   \\ \hline
1                  & 0.54 (0.70)      & 0.67 (0.80)   & 0.46 (0.63)   \\
2                  & 0.51 (0.68)      & 0.61 (0.76)   & 0.41 (0.58)   \\
3                  & 0.55 (0.71)     & 0.55 (0.71)   & 0.44 (0.61)   \\
4                  & 0.54 (0.70)     & 0.63 (0.77)   & 0.44 (0.61)   \\
5                  & 0.50 (0.67)        & 0.65 (0.79)   & 0.43 (0.60)   \\ \hline\hline
$m$            & 0.53 (0.69)       & 0.62 (0.77)   & 0.44 (0.61)   \\
$\sigma$ & 0.02 (0.02)        & 0.05 (0.04) & 0.02 (0.02)
\end{tabular}
\end{table*}

Using U-Nets, for each random fold, the network is trained using the 1800
training images. After training, the average Jaccard Index score is obtained for
the remaining 200 test images. This is done for both the U-Net algorithm without
pre-/post-processing (Algorithm 1A) and the more complex algorithm with
pre-/post-processing (Algorithm 1B).

Examining the Table~\ref{tab:table}, it is clear that the preprocessing steps
(performing histogram equalization, etc.) significantly improved the
segmentation results. The average Jaccard Index for Algorithm 1A was 53\% with a
standard deviation of 2\% while the average Jaccard Index for Algorithm 1B was
62\% with a standard deviation of 4.7\%.

Comparing the U-Net results with state of the art neural networks while it is
lower than the top performers at the 2017 ISBI competition, the results were
similar to many other submissions using deep networks. For example, the top ISIC
2017 challenge submission, Yuan et al. \cite{10} who used fully
convolutional-deconvolutional networks obtained an average Jaccard Index of
0.765. Similarly, Berseth \cite{17} who also used a U-Net obtained the second
place in the 2017 challenge with a Jaccard Index of 0.762. There were also many
challenge submissions that received a Jaccard Index between 0.444 and 0.665,
some of which also used deep neural networks.

The biggest advantage using the proposed algorithm is the number of training
iterations required. For example, Yuan et al. trained with 500 epochs on the
2,000 images. Similarly, Berseth trained 200 epochs using 20,000 images (after
data augmentation). This means that the two networks above required 1,000,000
and 4,000,000 training iterations, respectively.

In comparison, the networks presented in this paper was trained with only 10,000
training iterations (just over 5 epochs of 1800 images). This means that the
network only required 1\% of the number of training iterations compared with the
other deep network methods. A low number of training iterations is used because
the algorithms presented in this paper is designed to work when only low
computational power is available (computers without GPUs).

Figure~\ref{fig:results} shows a comparison of the skin lesion segmentation
algorithms for a few select skin lesion images. The first column in the figure
shows the input RGB image. The second column contains the Ground Truth (binary
mask manually segmented by a dermatologist). Finally, the third and fourth
columns show the output of the U-Net trained with algorithms 1A and 1B
respectively. As is evident from the figure, while both algorithm 1A and 1B are
able to determine the location of the lesion, algorithm 1B is better at
determining the border of the skin lesion since algorithm 1A produced a smaller
image.

The last three rows show some pathological cases where one or both of the
algorithms are unable to properly segment the image. This includes the skin
lesion being very small, the presence of another object such as medical gauze or
sleeve, and the skin lesion being extremely faint. Looking at the pathological
cases, it is clear that the algorithm that performs histogram equalization
performs better in most circumstances.

Given that the clustering based approach was unsupervised and did not contain
any ``training'' components, it was not necessary to conduct cross-validation
over multiple folds. The output was always going to be the same for every image.
However, for the purposes of comparison against the U-nets based approach, the
values were calculated for each of the folds used in the earlier results. The
Jaccard Index calculated can be found summarized in the Table~\ref{tab:table},
last column. The Jaccard Index for the entire data set was calculated to be
0.443. This value is quite low in comparison to the values obtained using
U-nets. However, it is important to remember that this algorithm is much simpler
and quicker to execute and does not require expensive training.

Figure~\ref{fig:results} contains a visualization from the clustering algorithm
alongside the output from U-Nets for various image types. Clearly the clustering
results are good for images in which no external disturbance exists and color
variation is minimized. However, the algorithm performs poorly when artifacts
are present. For example, the algorithm incorrectly identifies the black border,
the medical gauze, and other dark objects as the skin lesion.

When comparing this with other submissions in the ISIC challenge, this approach
would be on par with the 21st place submission. Amongst the competition entries,
only one submission followed a clustering based approach submitted by Alvarez et
al. \cite{18}. Their solution produced a significantly better result with a
Jaccard Index of approximately 0.679. The main reason for this is that their
approach uses a much more complex technique to classify the clusters involving
an ensemble method which leveraged random forests and support vector machines.
When comparing the results from both approaches it is clear by virtue of the
Jaccard Index that the U-Net based approach (with preprocessing) is superior.

\section{Conclusion and Future Work} \label{sec:conclusion}
In conclusion, this paper presented an effective approach to automatically
segmenting skin lesions particularly when computers with GPUs may not be available. We described two algorithms, one leveraging U-Nets and another using unsupervised clustering, and put forward effective pre-processing techniques for both approaches to improve the segmentation performance. It is noted that the solutions outlined in this paper are on par with the mid-tier submissions in the ISIC competition while utilizing significantly less training resources than many submissions.

After comparing the clustering and U-Net algorithms, it was clear that U-Nets
produce a much better segmentation result, especially using the histogram
equalization based pre-processing step. The network only required 1\% of the
number of training iterations compared with the other deep network methods. This
could be particularly beneficial when only low computational power is available
(computers without GPUs)

In the future, it would be interesting to find out if the histogram equalization
based pre-processing step also significantly improves segmentation performance
when the U-Net is trained over hundreds of epochs, instead of just one epoch,
with a more powerful GPU similar to the top ISIC 2017 challenge submissions.

\end{document}